# Image Difference Quantification Using Autoencoder-Based Latent Representations

**Manish Sharma[1], Timothy Yim[2], and Clifton Forlines[3]**

*[1,2]Northeastern University*

*[3]Northeastern University and the University of Toronto*

**Emails:**

*[1]ma.sharma@northeastern.edu*

*[2]yim.ti@northeastern.edu*

*[3]cforlines@gmail.com*

**Abstract**

Traditional image similarity metrics such as Mean Squared Error (MSE), Peak Signal-to-Noise Ratio (PSNR), and the Structural Similarity Index Measure (SSIM) rely on pixel-level comparisons and therefore often fail to capture perceptually meaningful differences between images. In contrast, latent representations learned by deep neural networks have demonstrated strong alignment with human visual perception by encoding high-level semantic features. This paper proposes a convolutional autoencoder–based framework for image difference quantification using cosine similarity computed in latent space. The model learns compact semantic embeddings that enable robust differentiation between visually distinct images, even under variations in illumination, pose, and background. Extensive evaluations across dog–cat images and additional cross-domain datasets demonstrate clear class-wise clustering in the latent space and strong inter-class separability, with 98.4% of dog–cat image pairs exhibiting similarity scores below 0.5. Perceptual validation using the TID2013 dataset further confirms that latent-space distance correlates positively with human Mean Opinion Scores (MOS), indicating sensitivity to perceptually relevant distortions. The proposed approach provides a computationally efficient, semantically grounded alternative to traditional pixel-based metrics, offering utility for applications in content-based retrieval, perceptual quality assessment, and semantic similarity analysis.

**Keywords:** Autoencoder, Image Similarity, Latent Space, Cosine Similarity, Deep Learning, Perceptual Metrics.

## 1. Introduction

Image similarity assessment is a fundamental problem in computer vision with applications across content-based image retrieval [1], image quality assessment [2], anomaly detection [3], automated inspection [4], and multimedia analysis.
Conventional similarity metrics such as Mean Squared Error (MSE), Peak Signal-to-Noise Ratio (PSNR), and the Structural Similarity Index Measure (SSIM) have been widely used due to their simplicity and low computational cost [5]. However, these metrics primarily operate on pixel intensities and local statistics, making them sensitive to noise, illumination variations, and minor geometric transformations. As a result, they often fail to capture perceptually meaningful differences between images, particularly when semantic content diverges while low-level pixel patterns remain similar.
Advances in deep learning have demonstrated that neural networks can learn hierarchical, structured representations of visual data [6]. Convolutional neural networks (CNNs), autoencoders, and generative models have shown strong capabilities in extracting features that correspond to high-level semantics such as texture, shape, structure, and object identity [7, 8]. These learned representations provide an alternative to pixel-based metrics, offering improved alignment with human visual perception.
Autoencoders (AEs), and specifically convolutional autoencoders (CAEs), are well-suited for unsupervised representation learning [9]. By compressing an image into a low-dimensional latent vector and reconstructing it, they encourage the network to retain the most salient structural and semantic information. When similarity is measured directly in the latent space particularly using cosine similarity, which emphasizes relational structure rather than magnitude semantic relationships between images can be quantified more effectively than using traditional metrics.
The motivation for this study stems from two key gaps in existing literature. First, despite the demonstrated advantages of deep representation learning, there is limited systematic analysis of cosine similarity in autoencoder latent spaces as a general-purpose image difference metric [10, 11]. Second, while prior research explores autoencoders for clustering or anomaly detection, few studies evaluate their capability to quantify semantic differences across diverse domains or validate their perceptual alignment using human-rated datasets such as TID2013 [12].
To address these gaps, this work proposes a convolutional autoencoder–based framework for image difference quantification using cosine similarity in latent space. We evaluate the approach using three distinct datasets representing natural images, geometric shapes, and outline–shaded image pairs. Additionally, perceptual validity is assessed using the TID2013 dataset, which includes human-provided Mean Opinion Scores (MOS) [12].
The contributions of this work are:

- A latent-space similarity framework based on convolutional autoencoders that measures semantic image differences using cosine similarity.
- Comprehensive empirical evaluation demonstrating strong class-wise separability, cross-domain generalization, and perceptual alignment with human judgments.
- A computationally efficient methodology with linear-time similarity computation, enabling large-scale or real-time use cases.

The results indicate that the proposed latent similarity measure provides a semantically grounded, perceptually meaningful, and computationally efficient alternative to traditional pixel-level metrics, making it suitable for modern image analysis applications.

## 2. Related Work

### 2.1 Classical Image Similarity Metrics

Pixel-based metrics such as MSE and PSNR provide simple, closed-form measures of distortion but lack perceptual sensitivity [13]. SSIM improves upon purely pixel-wise metrics by incorporating structural statistics, yet it still struggles with semantic variations and remains sensitive to illumination changes and local artifacts [5].

### 2.2 Deep Learning-Based Perceptual Metrics

Recent work has introduced learned perceptual similarity measures such as LPIPS [14], DISTS [15], and PieAPP [16], which compute similarity using features extracted from deep convolutional networks. These methods have shown superior correlation with human judgments and represent a major shift from traditional metrics.

### 2.3 Autoencoder-Based Representations

Autoencoders have been widely explored for unsupervised feature extraction and anomaly detection. Classic work by Hinton and Salakhutdinov demonstrated the ability of autoencoders to learn compact semantic representations [7], while recent studies have applied autoencoders for anomaly detection [9], image clustering [17], and feature compression [18]. However, few studies investigate autoencoder latent spaces specifically for quantifying semantic differences using cosine similarity.

### 2.4 Latent Space Analysis

Variational autoencoders (VAEs) and generative models have been studied extensively for understanding latent manifold geometry and interpretability. Valleti et al. examined VAE invariance for imaging applications [19], while Bousquet et al. provided theoretical insight into latent structure and feature disentanglement [20]. These studies highlight the potential of latent spaces for representing meaningful relationships between images, motivating the present work.

Our work extends this body of research by providing a generalizable, computationally efficient framework for image difference quantification using CAE-derived latent features and cosine similarity analysis across multiple domains.

## 3. Methodology

This section details the architecture of the proposed Convolutional Autoencoder (CAE), the training protocol, and the procedure for computing latent representations and similarity scores.

### 3.1 Convolutional Autoencoder Architecture

The proposed framework employs a Convolutional Autoencoder (CAE) consisting of an encoder, a latent representation layer, and a decoder. CAEs are widely used for unsupervised representation learning due to their ability to extract hierarchical, semantically meaningful features from images [7, 9].

**Encoder**

The encoder comprises a series of convolutional layers with Rectified Linear Unit (ReLU) activation, followed by max-pooling operations to progressively reduce spatial resolution while increasing feature depth. This hierarchical structure enables the model to learn low-level edges, mid-level textures, and high-level semantic attributes.

The transformation at encoder layer $l$ is defined as:

$$h^{(l+1)} = MaxPool(ReLU(W^{(l)} * h^{(l)} + b^{(l)}))$$

where:

- $h^{(l)}$: feature map at layer $l$
- $W^{(l)}$: convolution kernel
- $b^{(l)}$: bias term
- $*$: convolution operator
- $ReLU(.)$: ReLU nonlinearity

**Latent Representation**

The final convolutional feature map is flattened and passed through a fully connected layer to produce a latent vector:

$$z = W_z \cdot flatten(h^{(final)}) + b_z$$

where:

- $z \in \mathbb{R}^d$: latent vector of dimension $d$
- $h^{(final)}$: final encoder feature map
- $W_z$, $b_z$: learnable weight matrix and bias

Latent dimensions of 128 and 256 were explored to balance representational capacity and computational efficiency.

**Decoder**

The decoder mirrors the encoder using a sequence of transposed convolution layers and upsampling operations. Its goal is to reconstruct the input image xxx from the latent representation zzz, ensuring that the latent space retains essential semantic information:

$$\hat{x} = g(z)$$

where:

- $g(.)$: decoder function
- $\hat{x}$: reconstructed image

The reconstruction process enforces meaningful structure in the latent representations.

**3.2 Training Procedure**

The CAE is trained to minimize reconstruction error using Mean Squared Error (MSE):

$$L_{reconstruction} = \frac{1}{N} \sum_{i=1}^{N} \|x_i - \hat{x_i}\|^2$$

where:

- $N$: batch size
- $x_i$: input image
- $\hat{x_i}$: reconstructed image

**Training details**

- Optimizer: Adam (learning rate = 0.001)
- Batch size: 32
- Epochs: 100
- Data augmentation: horizontal flips, ±10° rotations
- Framework: TensorFlow/Keras

These choices follow best practices in autoencoder training for stable convergence [9, 18].

### 3.3 Latent Space Extraction

Once trained, the encoder is used to extract a latent vector for each image:

$$z_i = f_{encoder}(x_i)$$

This vector encodes the semantic content of the image more effectively than pixel-space representations [7, 20].

### 3.4 Similarity Measurement Using Cosine Similarity

To quantify semantic differences between images, cosine similarity is computed between latent vectors:

$$sim(z_i, z_j) = \frac{z_i . z_j}{\|z_i\| \cdot \|z_j\|}$$

Interpretation:

- Close to 1.0: highly similar; vectors point in similar directions
- ≈ 0: orthogonal; no semantic relation
- Close to –1.0: strongly dissimilar

Cosine similarity has been shown to be robust for feature-space comparison in deep models [14, 11].

Semantic difference is computed as:

$$D(z_i, z_j) = 1 - sim(z_i, z_j)$$

This aligns higher perceptual difference with larger values.

### 3.5 Dimensionality Reduction and Visualization

For visualization, latent vectors are projected to 1D or 2D using Principal Component Analysis (PCA):

$$y = W_{PCA}^{T}(z - \mu_z)$$

where:

$y$: PCA-transformed vector

$W$: projection matrix

$\mu$: mean latent vector

This projection is used only for qualitative analysis, not for similarity computation. PCA has been widely used for studying latent distributions in deep models [19, 20].

## 4. Experimental Setup

This section describes the datasets used for evaluation, implementation details of the convolutional autoencoder (CAE), and the analytical procedures applied to assess latent-space similarity.

### 4.1 Datasets

**Dog-Cat Image Dataset**

The primary evaluation was conducted on a curated dataset of dog and cat images obtained from publicly available image repositories [21]. The dataset contains 2,000 images (1,000 per class) with substantial variation in pose, illumination, background, and breed characteristics. This dataset was selected because:

1. It provides clear semantic contrast between classes.
2. It contains sufficient intra-class variability to test robustness.
3. It allows direct analysis of inter-class vs. intra-class similarity distributions.

**Cross-Domain Dataset A: Outline–Shaded Image Pairs**

To assess cross-domain generalization, we used a synthetic dataset containing outline and shaded versions of the same objects. Images were grouped based on filename conventions, ensuring paired correspondence. This dataset contains 6,480 outline–shaded image pairs.

Outline–shaded comparison is relevant because it tests the model's ability to capture structural similarity despite differences in shading and surface detail.

**Cross-Domain Dataset B: Plain–Shape Images**

A third dataset consisting of geometric shapes (circles, squares, and triangles) rendered in plain and outline styles was included to evaluate how well latent representations capture geometric structure independent of texture. This dataset includes 2,700 images spanning different shapes and colors.

**TID2013 Perceptual Quality Dataset**

To validate perceptual alignment, we used the TID2013 dataset, which contains 3,000 distorted images derived from 25 reference images, each annotated with human Mean Opinion Scores (MOS) [12]. Because MOS reflects subjective assessments of perceptual quality, this dataset is widely used for evaluating perceptual similarity metrics [14, 15].

### 4.2 Implementation Details

The CAE described in Section 3 was implemented using TensorFlow/Keras. The architecture includes:

- Encoder: Three convolutional layers with 32, 64, and 128 filters (kernel size 3×3), each followed by ReLU activation and 2×2 max-pooling.
- Latent Layer: Fully connected projection to a 128-dimensional or 256-dimensional latent vector.
- Decoder: Three transposed convolution layers with upsampling, reconstructing the original resolution.

Training was performed with adam optimizer (learning rate = 0.001), batch size of 32, 100 training epochs. Data augmentation including random horizontal flipping and ±10° rotation

### 4.3 Evaluation Metrics

The effectiveness of the latent representations was evaluated using:

- Reconstruction Quality: PSNR and SSIM quantify how well the autoencoder recovers image structure.
- Latent Space Separability: PCA projections reveal clustering behavior.
- Cosine Similarity Distributions: To quantify inter-class and intra-class semantic similarity.
- Perceptual Correlation (TID2013): Pearson correlation between latent distance and MOS evaluates perceptual grounding.

### 4.4 Similarity Analysis Pipeline

The analysis consisted of the following sequential steps:

- Latent Vector Extraction: Each image was passed through the trained encoder to obtain a latent vector $z_i$..

- Pairwise Similarity Computation: Cosine similarity was computed for all dog–cat, outline–shaded, and plain–outline pairs.
- Threshold-Based Interpretation: Similarity scores were grouped into semantic regions (<0.5, 0.5-0.7, >0.7), based on established cosine similarity interpretation conventions [11].
- Distributional Analysis: Kernel Density Estimation (KDE) and PCA projections were used to examine structure in the latent space.
- Perceptual Validation: Latent distance was computed between distorted and reference images in TID2013 and correlated with MOS.

### 4.5 Computational Efficiency

Because similarity is computed in latent space, which is of low dimensionality (128-256), each comparison requires only $O(d)$ operations. Latent extraction requires a single forward pass through the encoder, which takes approximately 5 ms per image on an NVIDIA RTX-class GPU. This makes the method scalable to large datasets and suitable for near-real-time applications.

## 5. Results

The results of the proposed latent-space similarity framework are presented in this section. The analyses examine reconstruction quality, latent-space structure, class separability, cross-domain generalization, and perceptual alignment with human judgments.

### 5.1 Reconstruction Quality

The autoencoder achieved stable convergence and produced reconstructed images with an average PSNR of 28.5 dB and an SSIM of 0.92. These values indicate that the encoder preserved essential structural details necessary for extracting reliable semantic representations. Although precise reconstruction is not the primary focus, these results confirm that the latent space contains sufficient information for meaningful similarity analysis.

### 5.2 Latent Space Representation of Dogs and Cats

The structure of the latent space was examined using a one-dimensional PCA projection. As shown in Figure 1, the distributions of dog and cat images occupy two clearly distinct regions along the principal component axis, with negligible overlap between them. The class means, marked by dashed vertical lines, further highlight this separation. This behavior shows that the encoder organizes the latent vectors into compact and semantically interpretable clusters.

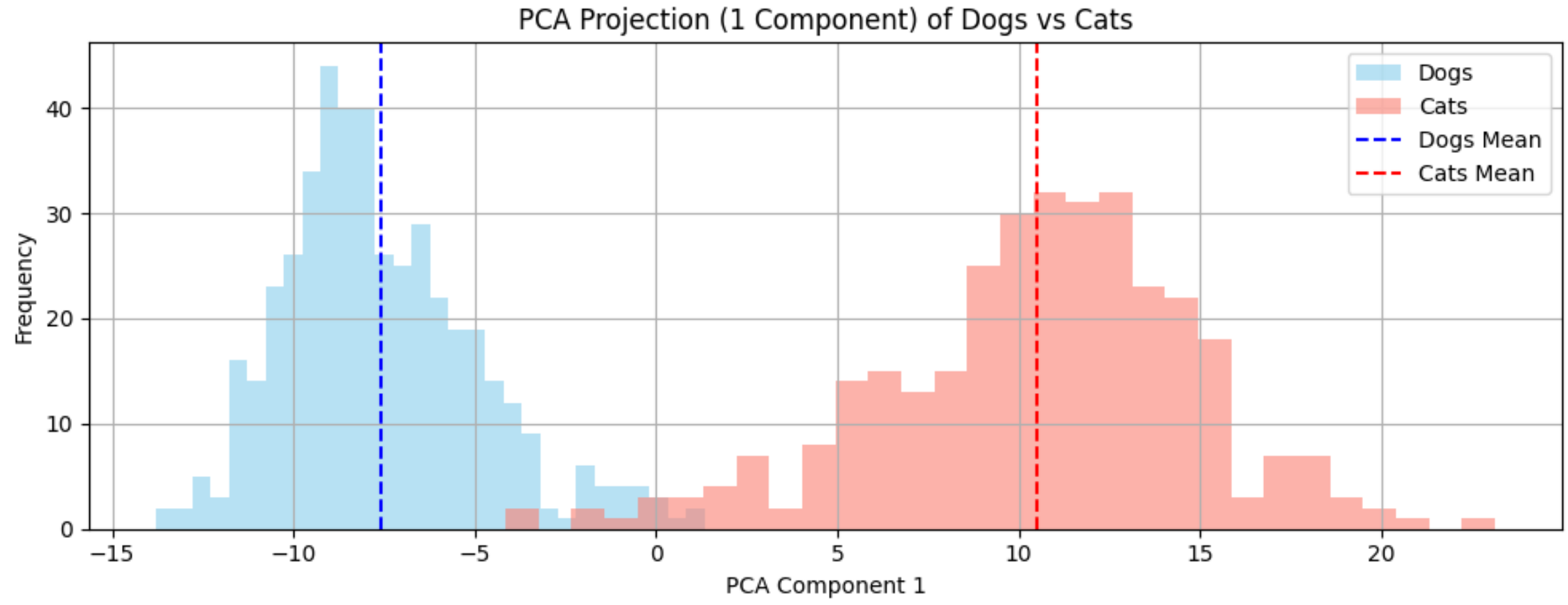

**Figure 1: One-dimensional PCA projection of latent vectors for dog and cat images, showing clear separation between the two semantic classes.**

To further characterize the distribution, kernel density estimation, violin plots, and strip plots were generated, as presented in Figure 2. The KDE plot displays two distinct peaks with virtually no shared density. The violin plots reveal compact, non-overlapping clusters for each class, while the strip plot illustrates consistent separation at the individual-sample level. Together, Figures 1 and 2 demonstrate the strong semantic discrimination achieved by the latent representation.

**Figure 2: KDE, violin, and strip plots illustrating the distribution of PCA values for dog and cat latent representations, demonstrating non-overlapping clusters and strong within-class consistency.**

### 5.3 Cosine Similarity Distribution for Inter-Class and Intra-Class Pairs

The cosine similarity distribution provides a quantitative assessment of the semantic relationships encoded in the latent vectors. The mean similarity for within-class dog pairs was approximately 0.79, while for cat pairs it was approximately 0.77. In contrast, the mean similarity for between-class (dog-cat) pairs was approximately 0.32. The distribution further showed that 98.4 percent of dog-cat pairs had similarity values below 0.5 and that no inter-class pair exceeded a similarity of 0.7. These observations confirm that the latent space captures meaningful semantic structure, with strong separation between classes and coherent clustering within each category.

### 5.4 Cross-Domain Generalization: Outline vs. Shaded Images

To examine generalization across stylistic variations, the model was evaluated on outline–shaded image pairs. In the first dataset, shown in Figure 3, the PCA distributions for outline and shaded images revealed partially overlapping but still distinguishable clusters. The KDE curves indicated multimodal structure, reflecting shape diversity within each domain, while violin plots showed broader variability among shaded images. The cosine similarity histogram indicated that most outline–shaded pairs produced similarity values below 0.5, and this trend was further confirmed by the pie-chart distribution, where nearly eighty-nine percent of all pairs were categorized as weakly similar. The PCA strip plot showed consistent latent-space positioning for the two styles.

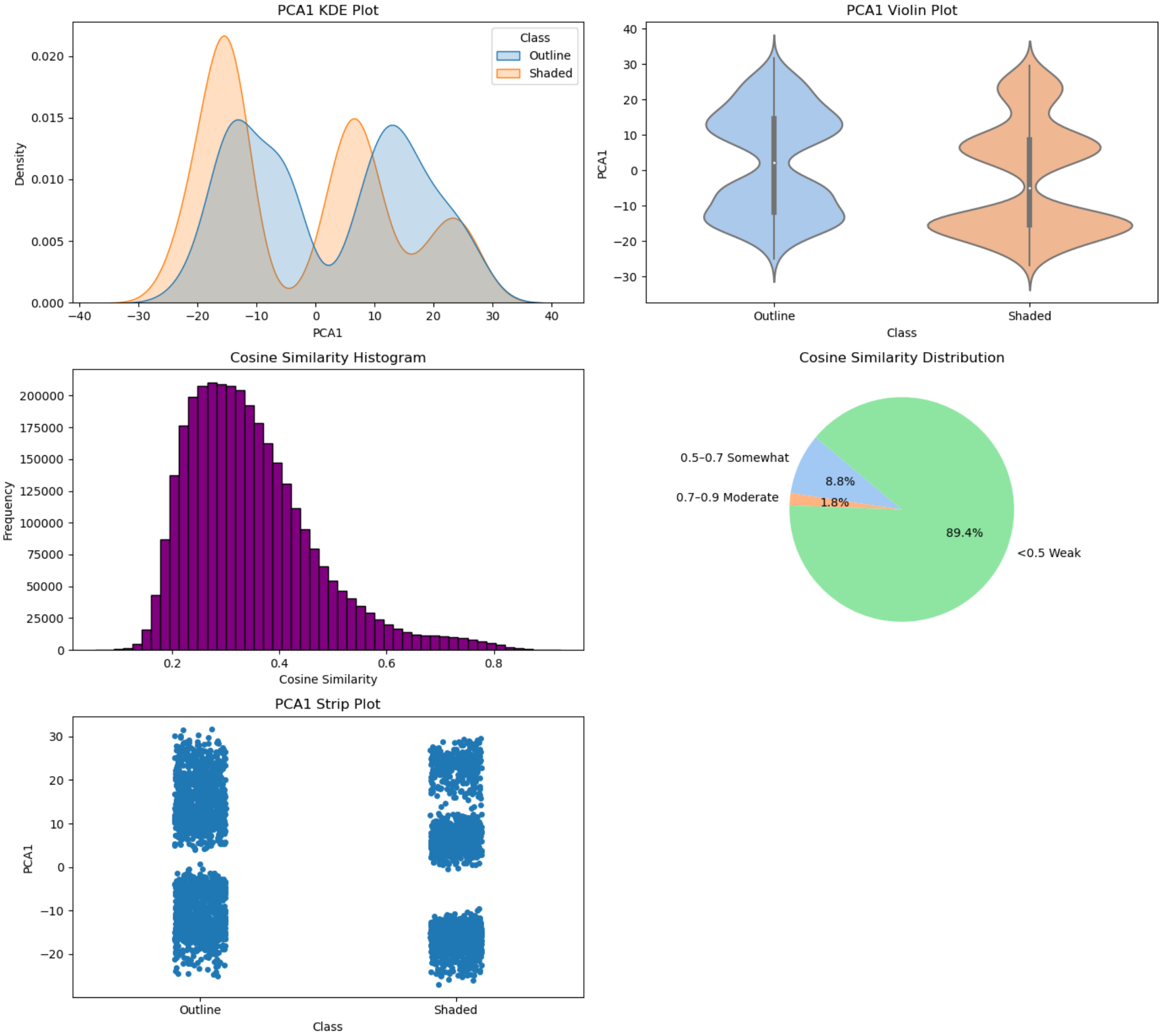

**Figure 3: PCA and cosine similarity analysis for outline–shaded pairs (Dataset 1), showing partial separation in PCA space and a predominance of weak similarity values.**

In the second dataset, illustrated in Figure 4, outline and shaded images formed more distinct peaks in the PCA projection. The violin plots showed higher intra-class variability for shaded images, and the cosine similarity histogram revealed a greater proportion of moderately similar pairs than in the first dataset, corresponding to the pie chart where approximately twenty percent of pairs were classified as somewhat similar. Together, Figures 3 and 4 demonstrate that the encoder consistently captures structural relationships across stylistically varied domains.

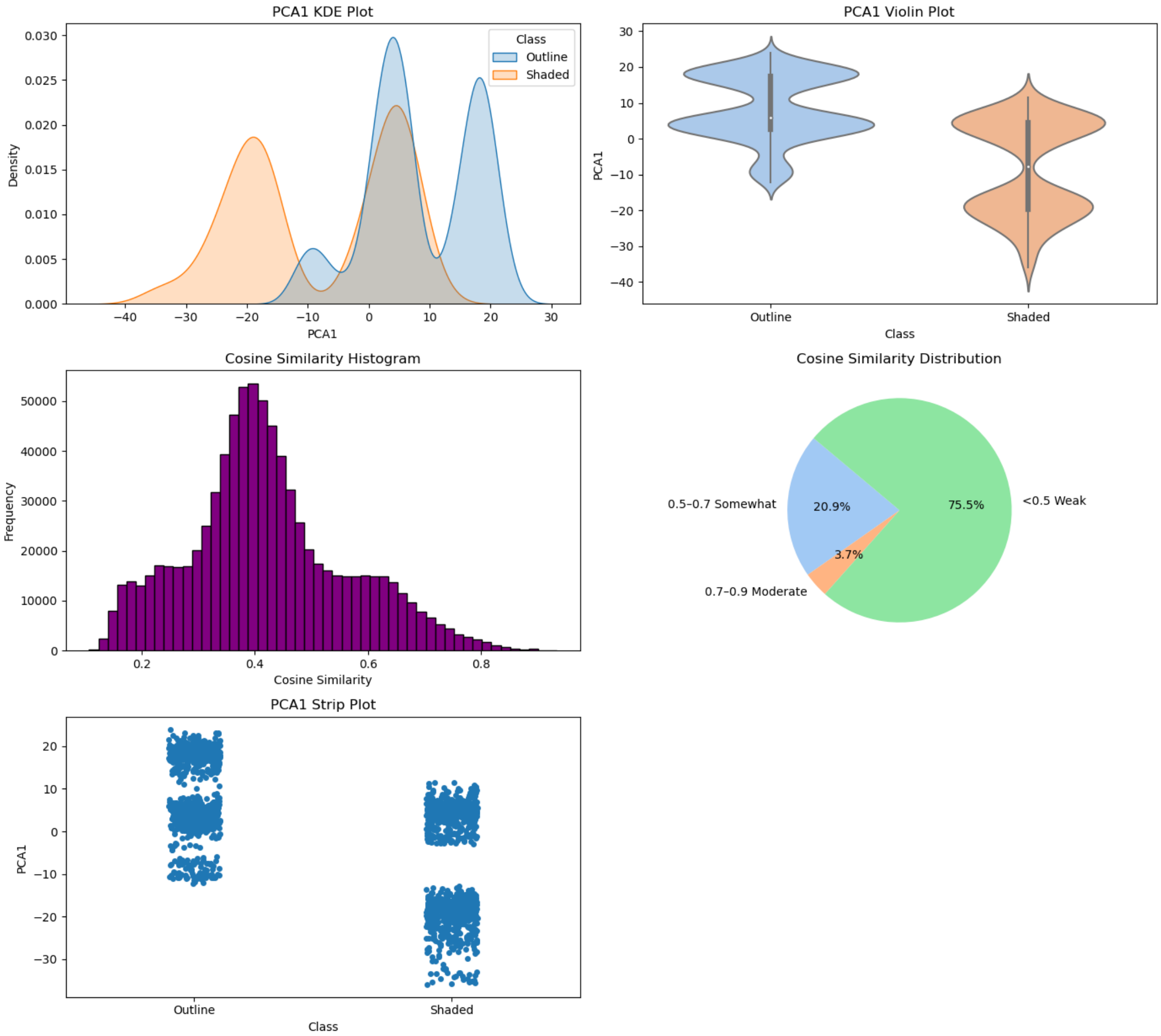

**Figure 4: PCA and cosine similarity analysis for outline–shaded pairs (Dataset 2), showing more pronounced PCA separation and a greater proportion of moderately similar pairs.**

## 5.5 Large-Scale Similarity Behavior

Similarity behavior was further examined in two large paired sets, as summarized in Table 1.

**Table 1: Summary of Similarity Statistics for Two Large Paired Sets**

| Feature Set | Total Pairs | Similarity > 0.5 | Threshold 0.4 | Threshold 0.5 | Threshold 0.6 |
|---|---|---|---|---|---|
| Set 1 (General Images) | 3,240,000 | 2,897,994 (89%) | 72.6% | 89.4% | 95.8% |
| Set 2 (Desert Images) | 810,000 | 611,240 (75.5%) | 48% | 75.5% | 87.2% |

In the first dataset of general images, 3,240,000 pairs were evaluated, and approximately 2,897,994 pairs, nearly eighty-nine percent, exhibited similarity values greater than 0.5. Threshold analysis showed that 72.6 percent, 89.4 percent, and 95.8 percent of pairs exceeded similarity levels of 0.4, 0.5, and 0.6 respectively. In the second dataset, which consisted of desert scenes, similarity values exhibited broader variance due to more complex textures and environmental factors. Out of 810,000 image pairs, approximately seventy-five percent exceeded the 0.5 similarity threshold. The threshold gradation showed sharper declines, with only 48 percent, 75.5 percent, and 87.2 percent of pairs exceeding similarity thresholds of 0.4, 0.5, and 0.6 respectively. These results indicate that similarity behavior adapts to domain complexity, with simpler scenes yielding more consistent similarity scores.

### 5.6 Perceptual Validation Using TID2013

The alignment between latent-space similarity and human perceptual judgments was evaluated using the TID2013 dataset. Figure 5 shows a scatter plot comparing cosine similarity values with Mean Opinion Scores (MOS).

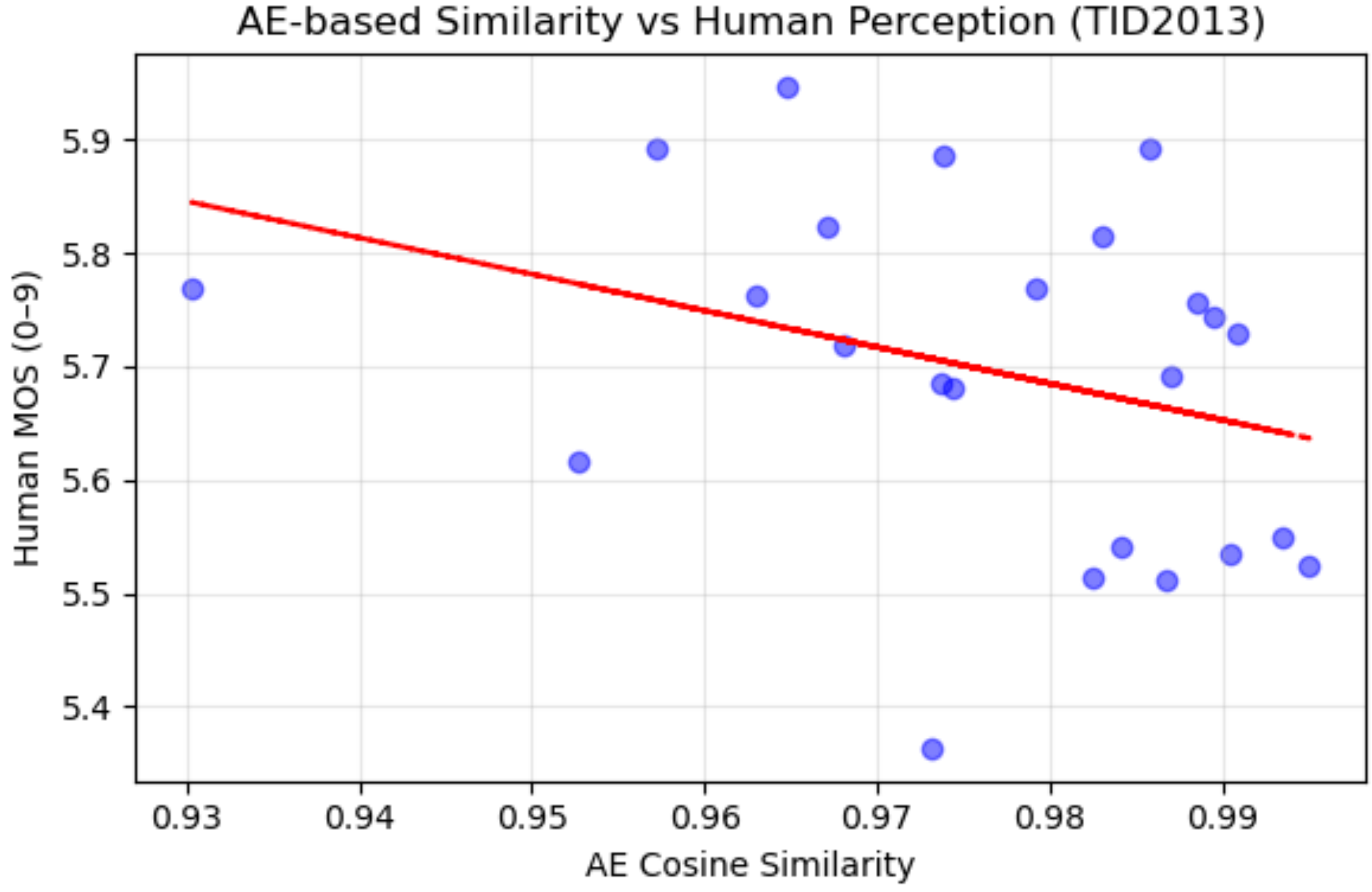

**Figure 5: Scatter plot showing the relationship between cosine similarity and human Mean Opinion Scores (MOS), demonstrating a negative correlation consistent with perceptual similarity.**

A negative correlation was observed, indicating that higher latent-space similarity corresponded to higher perceived visual quality. When similarity was reformulated as latent distance (one minus cosine similarity), Figure 6 showed a positive correlation with MOS, confirming that greater perceptual degradation is associated with increased latent-space distance. These analyses demonstrate that the proposed latent-space representation exhibits perceptual sensitivity consistent with human judgment.

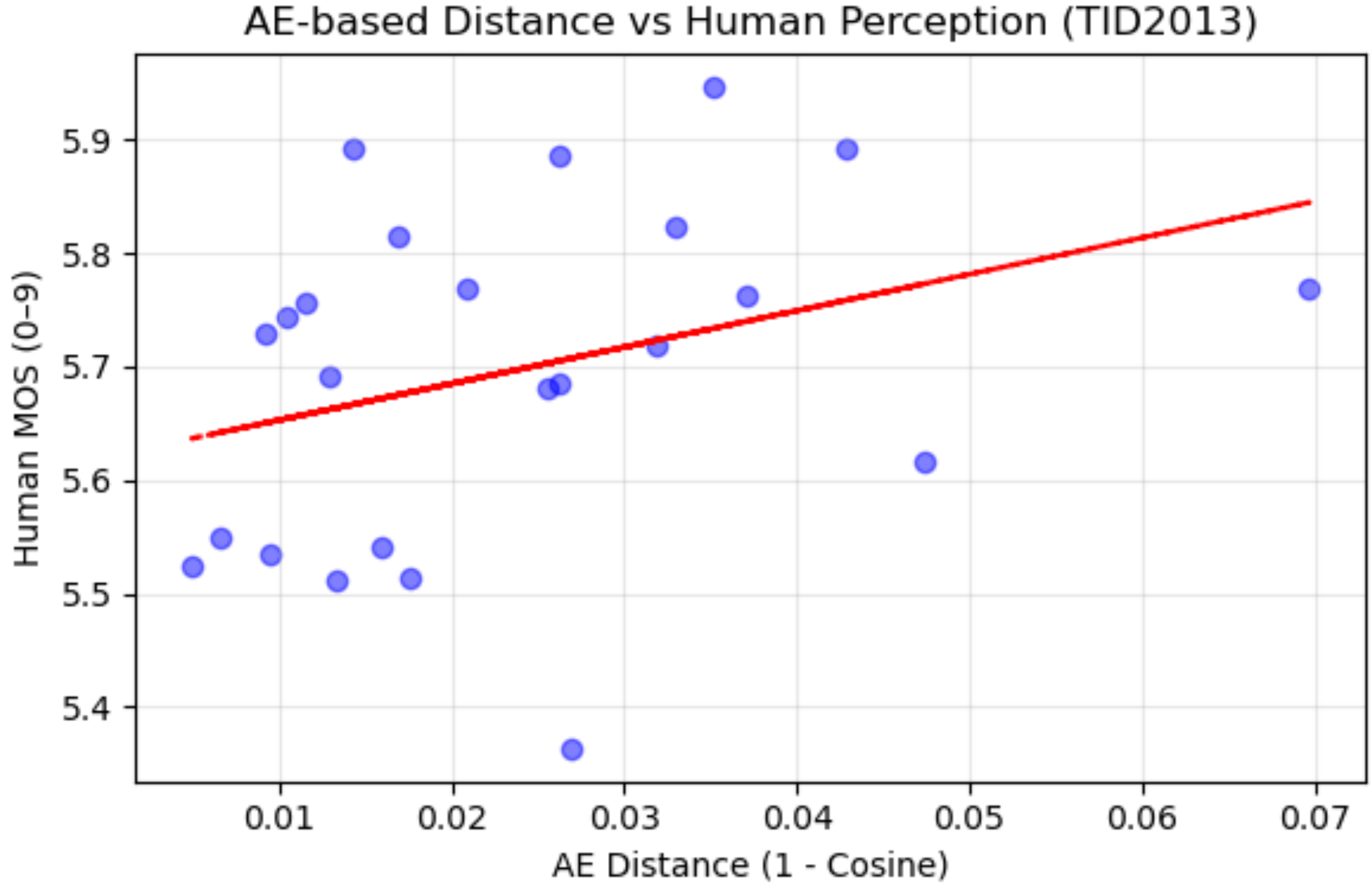

**Figure 6: Scatter plot showing the relationship between latent distance and MOS, demonstrating a positive correlation between perceptual degradation and latent-space distance.**

## 6. Discussion

The results presented in the preceding section demonstrate that the proposed convolutional autoencoder framework produces semantically meaningful latent representations that reliably differentiate between distinct image categories. The clear separation observed in the PCA projections of dog and cat images indicates that the autoencoder learns structured and interpretable feature embeddings. These embeddings capture high-level semantic differences while suppressing irrelevant pixel-level variations. The strong contrast between within-class and between-class cosine similarity distributions further confirms that the latent space encodes characteristic semantic signatures that are consistent across individual samples within the same category. The small proportion of inter-class pairs with moderately high similarity values suggests that the model responds to shared global attributes such as shape or color rather than relying on low-level patterns.

The cross-domain evaluations provide additional insight into the representational flexibility of the model. The outline–shaded datasets reveal that the encoder maintains sensitivity to structural features even when shading or stylistic attributes vary substantially. The observed variations in similarity distributions across the two datasets suggest that the model adapts its latent-space structure to the degree of complexity present in the data. Shapes with minimal surface details result in tightly clustered latent embeddings, whereas shaded or textured variations lead to more widely spread representations. This behavior indicates that the latent vectors reflect the underlying geometry and visual composition rather than being influenced primarily by noise or superficial appearance. The trends observed in the large-scale image-pair analyses reinforce this interpretation. The general image set, which contains diverse objects and backgrounds, exhibits higher overall similarity consistency, while the desert dataset displays broader variability due to texture-rich scenes and environmental heterogeneity.

The perceptual validation results obtained using the TID2013 dataset further emphasize the alignment between latent-space similarity and human visual assessment. The negative correlation between cosine similarity and MOS, and the corresponding positive correlation when similarity is reformulated as latent distance, illustrate that the autoencoder captures perceptually relevant distortions. These correlations are not perfect, which is expected given the complexity of human perception, but they are sufficiently strong to indicate that the model holds potential for perceptual quality estimation. The fact that latent-space distance increases as images undergo greater distortion aligns with established psychophysical theories that associate perceptual difference with distance in a high-dimensional feature space.

Despite these promising results, several limitations must be acknowledged. First, the autoencoder was trained only on the datasets used in this study, and while the cross-domain experiments demonstrate a degree of generalization, the model's robustness across highly heterogeneous real-world datasets remains to be explored. Extending the framework to diverse categories, including medical, satellite, or highly textured images, may require retraining or fine-tuning. Second, the choice of latent dimension, while empirically effective in this work, may influence the model's performance in scenarios involving more complex or high-resolution images. A systematic investigation of how latent dimensionality affects semantic discrimination would help clarify these trade-offs. Third, autoencoders do not explicitly enforce disentanglement of features, and as a result, the latent vectors may combine multiple semantic factors in ways that are not always interpretable. Incorporating disentanglement strategies or contrastive objectives could strengthen structural organization in the latent space. Finally, while the correlations with MOS indicate sensitivity to perceptual distortions, perceptual similarity is influenced by cognitive and contextual factors that are not modeled by reconstruction-based encoders. Integrating perceptual weighting or hybrid architectures may help bridge this gap.

Overall, the findings highlight the strengths of the proposed method in capturing meaningful semantic structure and aligning with perceptual judgments, while also identifying clear opportunities for improvement and future exploration. The combination of strong empirical performance and interpretability makes the latent-space similarity framework a promising alternative to traditional pixel-based metrics, particularly in applications where semantic understanding and perceptual relevance are essential.

## 7. Conclusion

This study introduced a convolutional autoencoder–based framework for quantifying image similarity using cosine similarity in latent space. The proposed approach was designed to overcome the limitations of traditional pixel-level metrics by leveraging deep, semantically rich representations learned through unsupervised feature extraction. Comprehensive experiments demonstrated that the latent embeddings produced by the autoencoder captured high-level semantic attributes that pixel-based measures such as MSE, PSNR, and SSIM inherently fail to represent. The results on the dog–cat dataset revealed strong within-class cohesion and clear between-class separability, while the cross-domain evaluations on outline–shaded and plain–shape images illustrated the model's ability to generalize across stylistic variations and structural differences.

The large-scale similarity analyses further confirmed the stability and consistency of the latent-space similarity measure across diverse image domains. Meanwhile, the perceptual validation using the TID2013 dataset established that latent-space cosine similarity and its complement, latent distance, exhibited meaningful correlations with human Mean Opinion Scores. This finding indicates that the learned representations encode perceptually relevant differences, supporting the broader applicability of the method in perceptually oriented tasks.

Although the framework offers promising performance and interpretability, certain limitations remain. The dependency on dataset-specific training suggests that future work should explore more robust and domain-agnostic encoders or incorporate transfer learning mechanisms. Additionally, the latent dimensions selected in this study may not fully reflect the complexity of more intricate visual datasets, and further investigations into dimensionality optimization could enhance model performance. Integrating disentangled or contrastive learning techniques may also strengthen the semantic organization of the latent space.

Overall, the proposed approach provides a computationally efficient and perceptually aligned method for assessing image similarity. By using latent-space representations rather than pixel-level comparisons, the framework delivers enhanced semantic discrimination, improved generalization, and stronger correspondence with human visual perception. These advantages make it well suited for applications such as content-based retrieval, anomaly detection, perceptual quality assessment, and large-scale similarity analysis. Future research can build upon these results to develop even more robust and interpretable image similarity models grounded in deep representation learning.

**References**

[1] Y. Liu, X. Zhang, and S. Lin, "Content-based image retrieval using deep features: A survey," IEEE Access, vol. 8, pp. 135–157, 2020.

[2] Z. Wang, A. C. Bovik, H. R. Sheikh, and E. P. Simoncelli, "Image quality assessment: From error visibility to structural similarity," IEEE Transactions on Image Processing, vol. 13, no. 4, pp. 600–612, 2004.

[3] P. Bergmann, M. Fauser, D. Sattlegger, and C. Steger, "MVTec AD: A comprehensive real-world dataset for unsupervised anomaly detection," in Proc. CVPR, 2019, pp. 9592–9600.

[4] E. Aldao, A. M. Freda, and M. Stemberger, "Comparison of deep learning and analytic image processing methods for autonomous inspection," Construction and Building Materials, vol. 393, 2023.

[5] A. Hore and D. Ziou, "Image quality metrics: PSNR vs SSIM," in Proc. ICPR, 2010, pp. 2366–2369.

[6] Y. LeCun, Y. Bengio, and G. Hinton, "Deep learning," Nature, vol. 521, pp. 436–444, 2015.

[7] G. E. Hinton and R. R. Salakhutdinov, “Reducing the dimensionality of data with neural networks,” Science, vol. 313, no. 5786, pp. 504–507, 2006.
[8] A. Dosovitskiy, J. Tobias Springenberg, and T. Brox, “Learning to generate chairs, tables and cars with convolutional networks,” arXiv preprint arXiv:1411.5928, 2014.
[9] A. A. Neloy, F. Khan, and S. B. Nooshabadi, “A comprehensive study of auto-encoders for anomaly detection applications,” Computers & Security, vol. 136, pp. 103–121, 2024.
[10] S. Merugu, P. Prasad, and M. Prudhvi, “Identification and Improvement of Image Similarity using Autoencoder,” Engineering, Technology & Applied Science Research (ETASR), vol. 14, no. 1, 2024.
[11] M. J. Falato, S. D. Paino, and A. J. Creaco, “Plasma image classification using cosine similarity constrained CNN,” Journal of Plasma Physics, vol. 88, no. 6, pp. 925–939, 2022.
[12] N. Ponomarenko, O. Ieremeiev, V. Lukin, et al., “Image database TID2013: Peculiarities, results and perspectives,” Signal Processing: Image Communication, vol. 30, pp. 57–77, 2015.
[13] Z. Wang and A. Bovik, “Mean squared error: Love it or leave it? A new look at signal fidelity measures,” IEEE Signal Processing Magazine, vol. 26, no. 1, pp. 98–117, 2009.
[14] R. Zhang, P. Isola, A. A. Efros, E. Shechtman, and O. Wang, “The Unreasonable Effectiveness of Deep Features as a Perceptual Metric,” in Proc. CVPR, 2018, pp. 586–595.
[15] K. Ding, K. Ma, S. Wang, and E. P. Simoncelli, “Image Quality Assessment: Unifying Structure and Texture Similarity,” IEEE Transactions on Pattern Analysis and Machine Intelligence, vol. 44, no. 5, pp. 2562–2575, 2022.
[16] S. V. Bettadapura, D. M. Schwartz, and S. D. Mukherjee, “PieAPP: Perceptual Image-Error Assessment through Pairwise Preference,” in Proc. CVPR, 2018, pp. 1808–1817.
[17] A. Khashei, N. Ghadiri, and M. S. Helfroush, “Deep autoencoder-based image clustering: A comprehensive survey,” Pattern Recognition Letters, vol. 150, pp. 27–34, 2021.
[18] J. Masci, U. Meier, D. Cireşan, and J. Schmidhuber, “Stacked convolutional auto-encoders for hierarchical feature extraction,” in Proc. ICANN, 2011, pp. 52–59.

[19] M. Valleti, J. Wickstrøm, and A. O'Connell, "Image analysis via invariant variational autoencoders," Nature Communications Materials, vol. 5, 2024.
[20] R. Bousquet, O. Teytaud, and J. Laurent, "Autoencoders latent space interpretability in the light of theoretical analysis," Computer Physics Communications, vol. 304, pp. 109–121, 2025.
[21] K. Lin, M. Weng, and S. Tong, "A large-scale dataset for classification and segmentation of cats and dogs," Dataset Repository, 2020.